\documentclass{article}

\usepackage{array}
\newcolumntype{C}[1]{>{\centering\let\newline\\\arraybackslash\hspace{0pt}}m{#1}}

\usepackage{microtype}
\usepackage{graphicx}
\usepackage{subfigure}
\usepackage{booktabs} 
\usepackage{amsfonts, amssymb}
\usepackage{fancybox, graphicx}
\usepackage{listings}

\usepackage{tikz, pgfplots}
\usepackage{amsmath, amsfonts, amssymb}

\usepackage[yyyymmdd]{datetime}

\usepackage{color} 
\usepackage{graphicx} 

\newtheorem{defn}{Definition}

\newtheorem{thm}{Theorem}

\usepackage{tikz}
\usetikzlibrary{shapes, arrows, positioning, decorations.markings}
\usetikzlibrary{calc}

\usetikzlibrary{fadings}

\tikzfading[name=fade out, inner color=transparent!0, outer color=transparent!100]

\definecolor {processblue}{cmyk}{0.96,0,0,0}
\tikzstyle{int}=[draw, fill=blue!20, minimum size=2em]
\tikzstyle{init} = [pin edge={to-,thin,black}]

\usetikzlibrary{shapes,shapes.multipart}
\usetikzlibrary{positioning,calc}  

\usetikzlibrary{shapes}
\usetikzlibrary{fit}
\usetikzlibrary{chains}
\usetikzlibrary{arrows}

\tikzstyle{plate} = [draw, rectangle, rounded corners, fit=#1]
\tikzstyle{wrap} = [inner sep=0pt, fit=#1]

\tikzstyle{caption} = [node distance=0] %
\tikzstyle{bottom plate caption} = [caption, node distance=0, inner sep=0pt,
below left=-5pt and 0pt of #1.south east] %

\tikzstyle{top plate caption} = [caption, node distance=0, inner sep=0pt,
below left=0pt and 0pt of #1.north east] %

\usetikzlibrary{positioning}

\tikzstyle{inputNode}=[draw,circle,minimum size=10pt,inner sep=0pt, top color =white , bottom color = processblue!20 ,
draw, processblue , text=blue]
\tikzstyle{stateTransition}=[-stealth, thick]

    \pgfplotsset{
    standard/.style={
        axis x line=middle,
        axis y line=middle,
        enlarge x limits=0.15,
        enlarge y limits=0.15,
        width=8cm,
        height=6cm,
        every axis x label/.style={at={(current axis.right of origin)},anchor=north west},
        every axis y label/.style={at={(current axis.above origin)},anchor=north east},
        every axis plot post/.style={mark options={fill=white}}
        }
    }

\usepackage{hyperref}

\usepackage[accepted]{icml2019}

\usepackage{tikz}
\usetikzlibrary{shapes, arrows, positioning, decorations.markings}

\definecolor {processblue}{cmyk}{0.96,0,0,0}
\tikzstyle{int}=[draw, fill=blue!20, minimum size=2em]
\tikzstyle{init} = [pin edge={to-,thin,black}]

\usetikzlibrary{shapes}
\usetikzlibrary{fit}
\usetikzlibrary{chains}
\usetikzlibrary{arrows}

\begin{document}

\twocolumn[
\icmltitle{Understanding Feature Selection and Feature Memorization in  Recurrent Neural Networks}



\icmlsetsymbol{equal}{*}

\begin{icmlauthorlist}

\icmlauthor{Bokang Zhu}{buaa}
\icmlauthor{Yongyi Mao}{goo}
\icmlauthor{Richong Zhang}{buaa}
\icmlauthor{Dingkun Long}{buaa}
\end{icmlauthorlist}

\icmlaffiliation{buaa}{BDBC and SKLSDE, School of Computer Science and Engineering, Beihang University, 37 Xueyuan Road, Beijing 100191, China}
\icmlaffiliation{goo}{School of Electrical Engineering and Computer Science, University of Ottawa, 800 King Edward Avenue, K1N 6N5 ON, Canada}

\icmlcorrespondingauthor{Yongyi Mao}{yymao@eecs.uottawa.ca}

\icmlkeywords{Machine Learning, ICML}

\vskip 0.3in
]



\printAffiliationsAndNotice{\icmlEqualContribution} 

\begin{abstract}
	In this paper, we propose a test, called Flagged-1-Bit (F1B) test, to study the intrinsic capability of recurrent neural networks in sequence learning. 
Four different recurrent network models are studied 
both analytically and experimentally using this test.  Our results suggest that in general there exists a conflict between feature selection and feature memorization in sequence learning with recurrent neural networks. 
Such a conflict can be resolved either using a gating mechanism as in LSTM or by increasing the state dimension as in Vanilla RNN. 
Gated models resolve this conflict by adaptively adjusting their state-update equations, whereas Vanilla RNN resolves this conflict by assigning different dimensions different tasks. 
Insights into feature selection and memorization in recurrent networks are given. 

\end{abstract}

\section{Introduction}

Over the past decade, the revival of neural networks has revolutionized the field of machine learning, in which deep learning\cite{lecun2015deep} now prevails. Despite the stunning power demonstrated by the deep neural networks,  fundamental understanding concerning how these networks work remains nearly vacuous. As a consequence, designing deep network models in practice primarily relies on experience, heuristics and intuition. At present, deep neural networks are still regarded as ``black boxes''. Understanding and theorizing the working of these networks are seen, by many, as the utmost important topic in the research of deep learning\cite{informationBottleNeck}.

Among various frameworks in deep learning, recurrent neural networks (RNNs) \cite{Elman1990FindingSIRNN, hochreiter1997longLSTM, Cho2014OnTPGRU} have been shown particularly suited for modelling sequences, natural languages,  and time-series data\cite{sundermeyer2012lstm, mikolov2010recurrent, wu2016investigating}.  Their success in sequence learning has been demonstrated in a large variety of practical applications
\cite{Cho2014Learning,kalchbrenner2016neural,Cho2014Learning,cheng-lapata:2016:P16-1,cao2015ranking,nallapati2016abstractive,wen2015semantically,lewis2017deal,geiger2014robust,zeyer2017comprehensive}. But, in line with the overall development of deep learning, very limited understanding is available to date regarding how RNNs really work. For example, the following questions yet await for answers:  How does an RNN effectively detect and continuously extract the relevant temporal structure as the input sequence is feeding through the network? How does the network maintain the extracted feature in its memory so that it survives
in the incoming stream of noise or disturbances? Investigating these questions is the primary interest of this paper.

In this work, we construct a particular test, referred to the ``Flagged-1-Bit'' (F1B) test to answer these questions. In this test, we feed a
random sequence of ``bit pairs'', $(X_t^{\rm I}, X_t^{\rm F})$ to a recurrent network. The $X^{\rm F}$-components of the sequence contain a single flag ($+1$) at a random time $L$; the $X^{\rm I}$-components of the sequence are independent bits ($\pm 1$'s), among which only the value at time $L$ carries the desired feature. The objective of the test is to check if an RNN, {\em irrespective of the training algorithm},  is able to {\em select} the feature  and {\em memorize} it until the entire bit-pair sequence is fed to the network.

 The F1B test has obviously a simplified setting.
  However, the feature selection and memorization challenges designed in this test are important aspects of learning with RNN. Specifically, detecting the desired temporal structure and maintaining it throughout the sequence duration must both be carried out through effectively using the network's size-limited memory, or dimension-limited state. Such challenges arguably exist nearly universally.

We study the behaviour of several RNNs in this test. The studied models include Vanilla RNN\cite{Elman1990FindingSIRNN}, LSTM\cite{hochreiter1997longLSTM}, GRU\cite{Cho2014OnTPGRU} and PRU\cite{long2016Prototypicalru}. We note that among these models, except for Vanilla RNN, all other models exploit some ``gating mechanism''. 

In this paper, we prove that all studied gated RNNs are capable of passing the test at state dimension $1$, whereas Vanilla RNN fails the
test at state dimension $1$ but passes it at state dimension $2$. Our theoretical analysis and experimental study reveal that there is in general a conflict between selecting and memorizing features in sequence learning. We show that such a conflict can be resolved by introducing gates in RNNs, which allows the RNN to  dynamically switch between selection and memorization. Without using gates, e.g., in Vanilla RNN, such a conflict can only be resolved by increasing the state dimension. Specifically, we show that with adequate state space, Vanilla RNN is capable of splitting feature selection and feature memorization, and assigning the two tasks each to a different state dimension. 

To the best of our knowledge, this work is the first to formulate the conflict between feature selection and feature memorization in sequence modelling. It is also the first rigorous analysis of RNN's capability in resolving this conflict. Although the F1B test has a rather simple structure, the microscopic insights we obtain concerning the behaviour of gated and non-gated RNNs can arguably be extrapolated to more complex settings of sequence learning.



%
In this paper, in addition to presenting the key results, we also make an effort to give some insights in the problem scope. Due to space limitation, detailed proofs, some further discussions and additional results are presented in Supplementary Material, to which we sincerely invite the reader.

\section{Recurrent Neural Networks}

A generic {\em recurrent neural network (RNN)} is essentially a dynamic system specified via 
the following two equations.
\begin{eqnarray}
\label{eq:stateUpdate}
s_t & = & F(s_{t-1}, x_t)\\
\label{eq:output}
y_t &= & G(s_t, x_t)
\end{eqnarray}
where $t$ denotes discrete time, $x_t$, $y_t$ and $s_t$ are respectively the {\em input}, {\em output} and {\em state} variables. Note that here all these variables can be in general vectors.

In machine learning, the use of RNN is primarily for learning temporal features in the input sequence $x_1, x_2, \ldots, x_n$. The extraction of temporal information is expected to be carried out by the {\em state-update equation}, or, (\ref{eq:stateUpdate}), where the state $s_t$ at time $t$ serves as a complete summary of the ``useful'' information contained in the past inputs up to time $t$. The network uses  (\ref{eq:output}), or the {\em output equation},  to generate output if needed.  

In this paper, since we are only concerned with how temporal features are extracted and maintained in the state variable, we largely ignore the exact form of the output equation. That is, we regard each recurrent network as being completely defined by its {\em state-update function} $F$ in  (\ref{eq:stateUpdate}). We use $\Theta$ to denote the parameter of function $F$.

In this setting an {\em RNN model} ${\cal F}$ is a family of state-update functions having the same parameterization. We now give a concise description of several RNN models, all of which are studied in this work.




\noindent{\bf Vanilla RNN}, denoted by ${\cal F}_{\rm RNN}$,  is among the earliest neural network models. It is conventionally known as ``recurrent neural network'' or simply ``RNN''\cite{Elman1990FindingSIRNN}. In the modern literature, it is referred to as Vanilla RNN. Its state-update equation is given by (\ref{eq:rnnStateUpdate}). 
\begin{eqnarray}
\label{eq:rnnStateUpdate}
s_t & = & \tanh(Wx_t + Us_{t-1} + b)
\end{eqnarray}


\noindent{\bf LSTM}, 
denoted by ${\cal F}_{\rm LSTM}$, was first presented in \cite{hochreiter1997longLSTM} and \cite{gers2000recurrent}. In this paper, we adopt the formalism of LSTM in \cite{gers2000recurrent}. Specifically,
to better control the gradient signal in back propagation, LSTM introduces several ``gating'' mechanisms to control its state-update equation, via (\ref{eq:LSTM_StateUpdate}) to (\ref{eq:LSTM_anotherOutput}).
{
\begin{eqnarray}
\label{eq:LSTM_StateUpdate}
i_t & = & \sigma(W^{\rm i}x_t + U^{\rm i}d_{t-1} + V^{\rm i}c_{t-1} + b^{\rm i})\\
f_t & = & \sigma(W^{\rm f}x_t + U^{\rm f}d_{t-1} + V^{\rm f}c_{t-1} + b^{\rm f})\\
o_t & = & \sigma(W^{\rm o}x_t + U^{\rm o}d_{t-1} + V^{\rm o}c_{t-1} + b^{\rm o})\\
\widetilde{c}_t & = & \tanh(W^{\widetilde{\rm c}}x_t + U^{\widetilde{\rm c}}d_{t-1} + V^{\widetilde{\rm c}}c_{t-1} +  b^{\widetilde{\rm c}})\\
c_t & = & i_t \odot \widetilde{c}_t + f_t \odot c_{t-1}
\label{eq:LSTM_gatedCombine}
\\
d_t & = & o_t \odot \tanh(c_t)
\label{eq:LSTM_anotherOutput}
\end{eqnarray}
}
Note that we use $\sigma(\cdot)$ to denote the sigmoid (i.e. logistic) function and $\odot$ to denote element-wise vector product. In the context of neural networks,
gating usually refers to using a dynamically computed variable in $(0, 1)$ to scale a signal, as in  (\ref{eq:LSTM_gatedCombine}) and (\ref{eq:LSTM_anotherOutput}).

In LSTM, the pair $(c_t, d_t)$ of variables together serve as the state variable $s_t$, if one is to respect the definition of state variable in system theory\footnote{In system theory, a state variable $s_t$ is a variable (or a set of variable), given which the future behaviour of the system is independent of its past behaviour. }.  However, in the literature of LSTM, it is customary to consider $c_t$ as the state variable, since in the design of LSTM \cite{hochreiter1997longLSTM},  $c_t$ is meant to serve as a memory, maintaining the extracted feature and carrying it over to the next time instant.  We will also take this latter convention in this paper. 

\noindent{\bf GRU}, denoted by ${\cal F}_{\rm GRU}$,  was presented in \cite{Cho2014OnTPGRU}. It uses a simpler gating structure compared to LSTM. Its state-update equation is given by (\ref{eq:GRU_StateUpdate}) to (\ref{eq:GRU_gatedCombine}).
{
\begin{eqnarray}
\label{eq:GRU_StateUpdate}
r_t & = & \sigma(W^{\rm r}x_t + U^{\rm r}s_{t-1} + b^{\rm r})\\
z_t & = & \sigma(W^{\rm z}x_t + U^{\rm z}s_{t-1} + b^{\rm z})
\label{eq:GRU_gate}
\\
\widetilde{s}_t & = & \tanh\left(W^{\widetilde{\rm s}}x_t + U^{\widetilde{\rm s}}(r_t \odot s_{t-1}) + b^{\widetilde{\rm s}}\right)\\
s_t & = & z_t \odot \widetilde{s}_t + (1-z_t)\odot s_{t-1}
\label{eq:GRU_gatedCombine}
\end{eqnarray}
}
Empirical results have shown that with this simplified gating structure, GRU can still perform comparatively or even better than LSTM \cite{chung2014empirical}. 


\noindent{\bf PRU}, or Prototypical Recurrent Unit, which we denote by ${\cal F}_{\rm PRU}$, is a recently presented model\cite{long2016Prototypicalru}. 
The state-update equation of PRU contains only a single gate as is given by (\ref{eq:PRU_StateUpdate}) to (\ref{eq:PRU_gatedCombine}).
\begin{eqnarray}
\label{eq:PRU_StateUpdate}
z_t & = & \sigma(W^{\rm z}x_t + U^{\rm z}s_{t-1} + b^{\rm z})\\
\widetilde{s}_t & = & \tanh(W^{\widetilde{\rm s}}x_t + U^{\widetilde{\rm s}}s_{t-1} + b^{\widetilde{\rm s}})\\
s_t & = & z_t \odot \widetilde{s}_t + (1-z_t)\odot s_{t-1}
\label{eq:PRU_gatedCombine}
\end{eqnarray}
It was shown in \cite{long2016Prototypicalru} that even with this minimum gate structure, PRU still performs very well. With the same total number of parameters, PRU was shown to  outperform GRU and LSTM. 

In summary, LSTM, GRU, and PRU are all ``gated models''. In particular, their state-update equations are all in a ``gated combination'' form, as in  (\ref{eq:LSTM_gatedCombine}), (\ref{eq:GRU_gatedCombine}), and (\ref{eq:PRU_gatedCombine}). Vanilla RNN is not gated.

\subsection{Related Works}

Despite their numerous applications, fundamental understanding of 
the working of RNN remains fairly limited.

In \cite{Siegelmann:1992}, Vanilla RNN is shown to be Turing-complete. In \cite{Bengio1994LearningLD,pascanu2013difficulty}, the difficulty of gradient-based training of Vanilla RNN is demonstrated and the gradient vanishing/exploding problem is explained. Significant research effort has been spent on curing this gradient problem, including, e.g., \cite{arjovsky2016unitary,dorobantu2016dizzyrnn,WisdomPHRA16}.  In \cite{CollinsSS16}, trainablity and information-theoretic capacities of RNNs are evaluated empirically. Recently \cite{ChenPS18} proposes a theory, based on the notion of dynamical isometry and statistical mechanics, to model signal propagation in RNN and to explain the effectiveness of gating.

\section{Flagged-1-Bit Test}

\begin{defn}[Flagged-1-Bit (F1B) Process]
A Flagged-1-Bit Process  $F1B(n)$ of length $n$  is a random sequence $X_1, X_2, \ldots, X_n$, defined as follows.
Each $X_t:=
\left[
\begin{array}{c}
X^{\rm I}_t\\
X^{\rm F}_t
\end{array}
\right]
$ is a length-2 vector, where both $X^{\rm I}_t$ and $X^{\rm F}_t$ take values in $\{\pm 1\}$. The two values $X^{\rm I}_t$ and $X^{\rm F}_t$
 are respectively referred to as an ``information bit'' and a ``flag bit''.   To generate the information bits, for each $t=1, 2, \ldots, n$, $X_t^{\rm I}$ is drawn independently from $\{\pm 1\}$ with equal probabilities. To generate the flag bits, first a random variable $L$ is drawn uniformly at random from the set of integers $\{1, 2, \ldots, n\}$, then $X_L^{\rm F}$ is set to $1$, 
 and $X_t^{F}$ is set to $-1$ for every 
 $t\neq L$.  
 \end{defn}

 In this process, we call $X^{\rm I}_L$ as the {\em flagged information bit}.  A sample path of $F1B(n)$ is called a {\em positive sample path} if $X^{\rm I}_L=+1$, and a {\em negative sample path} if $X^{\rm I}_L=-1$. 
 Table \ref{tab:F1B} shows two example sample paths of an $F1B(n)$ process. 

\begin{table}
{
\caption{\label{tab:F1B} Two sample paths of F1B(9). Top: positive, with $L=3$. Bottom:  negative, with $L=5$.}
\centerline{
\begin{tabular}{c}
\begin{tabular}{c|cccccccccc}
\hline
 $t$ & 1& 2& 3& 4& 5& 6& 7& 8&9\\ 
 \hline
 $x^{\rm I}_t$ & +1 & +1 & {\bf +1} & -1 & -1 &+1 &-1 &+1 &-1\\
 $x^{\rm F}_t$ & -1 & -1 & {\bf +1} & -1 & -1 &-1 &-1 &-1 &-1\\
 \hline
\end{tabular}
\\\hline
\hline
\begin{tabular}{c|cccccccccc}
 $t$ & 1& 2& 3& 4& 5& 6& 7& 8&9\\ 
 \hline
 $x^{\rm I}_t$ & +1 & -1 &  +1 & -1 & {\bf -1} &+1 &-1 &+1 &-1\\
 $x^{\rm F}_t$ & -1 & -1 & -1  & -1 & {\bf +1} &-1 &-1 &-1 &-1\\
 \hline
\end{tabular}
\end{tabular}
}
}
\end{table}

Now consider the learning problem with the objective of classifying the sample paths of $F1B(n)$ into the positive and negative classes. We wish to decide whether it is possible for an RNN model, at a given state dimension $K$,  to learn a perfect classifier. We hope this exercise gives some insights as to how RNN detects and remembers the desired features.  

%
To formally introduce the problem, first recall the definition of a binary linear classifier, where we will use ${\mathbb R}^K$ to denote the Euclidean space of dimension $K$ over the field ${\mathbb R}$ of real numbers: A binary {\em linear} classifier $C$ on ${\mathbb R}^K$ is a function mapping ${\mathbb R}^K$ to $\{\pm 1\}$ defined by 
{\small
\[
C(s)= 
\left\{
\begin{array}{ll}
+1, &{\rm if}~\beta^T s+ \gamma \ge 0.\\
-1, &{\rm otherwise}
\end{array}
\right.
\]}
where $\beta\in {\mathbb R}^K$ and $\gamma\in {\mathbb R}$ are parameters. We denote $(\beta, \gamma)$ collectively by $\Phi$.

For an arbitrary RNN with state-update function $F$ (and a fixed parameter setting), suppose that we draw a sample path $X=(X_1, X_2, \ldots, X_n)$ from $F1B(n)$ and feed it to the network $F$.  Suppose that $C_\Phi$ is a binary linear classifier,  parameterized by $\Phi$, on the state space of the network.
When $C_\Phi$ is applied to the state variable $S_n$, its error probability  $P_e(F, \Phi)$ is 
$P_e(F, \Phi) : = {\rm Pr}[C_\Phi(S_n) \neq X^{\rm I}_L]$,  
where the probability measure ${\rm Pr}[\cdot]$ is induced by the $F1B(n)$  process through the functions $F$ and $C_\Phi$. 


\begin{defn}[F1B Test]
Suppose that ${\cal F}$ is an RNN model with state dimension $K$ and parameterized by $\Theta$. The model ${\cal F}$ is said to pass the F1B test if for any integer $n>0$, there exist an initial configuration $s_0$ of $S_0$, a parameter setting $\Theta^*$ of $\Theta$, which defines $F_{\Theta^*}\in {\cal F}$,  and a setting $\Phi^*$ of $\Phi$ such that $P_e(F_{\Theta^*}, \Phi^*)=0$. 
\end{defn}

In essence, passing the F1B test means that for any given $n$, it is possible to find a parameter setting (depending on $n$) for the RNN model such that the positive and negative sample paths of the $F1B(n)$ process are perfectly and linearly separable at the final state of the RNN.  
This paper studies whether each afore-mentioned RNN model is capable of passing the test and at which state dimension. 


Although the proposed F1B test may seem a toy problem, it in fact captures fundamental challenges in sequence learning.
Specifically, 
for a model to pass this test, it needs to detect and select the flagged information bit when it enters the system, and load it (in some representation) in its state variable. Additionally, it must maintain this information, or ``memorize'' it, by preventing it from being over-written or suppressed by the subsequent noise sequence.  The test thus challenges the model in both its ability to select features and its ability to memorizing them.

For more discussion on the rationale of the F1B test, the reader is referred to Supplementary Material (Section 1).




\section{Theoretical Analysis}
\label{sec:theory}


\begin{thm}
\label{thm:K2KPlus1}
For any of the four models ${\cal F}_{\rm LSTM}$, ${\cal F}_{\rm GRU}$, ${\cal F}_{\rm PRU}$, ${\cal F}_{\rm RNN}$, if it passes the F1B test at state dimension $K$, it also passes the test at state dimension $K+1$.
\end{thm}

This result is intuitive, since increasing the state dimension should only increase the modelling power.  For each of the four models, the proof of this result relies on constructing a parameter $\Theta^{(K+1)}$ for the model with state dimension $K+1$ from a parameter $\Theta^{(K)}$, where $\Theta^{(K)}$ is a parameter that allows the model to pass the test at dimension $K$. The construction of $\Theta^{(K+1)}$ essentially makes the extra dimension in the model useless (i.e., equal to 0), thereby reducing the model to one with state dimension $K$ and parameter $\Theta^{(K)}$.

Due to this theorem, we only need to determine the smallest state dimension $K$ at which a model passes the test.

\subsection{Vanilla RNN}

\begin{thm}
At state dimension $K=1$, ${\cal F}_{\rm RNN}$ fails the F1B test.
\label{thm:RRN1d_fails}
\end{thm}
\begin{thm} At state dimension $K=2$, ${\cal F}_{\rm RNN}$ passes the F1B test. \label{thm:RNN2d_pass}
 \end{thm}
We note that in the proofs of these two theorems, we replace the $\tanh(\cdot)$ activation function in (\ref{eq:rnnStateUpdate}) with a piece-wise linear function $g(\cdot)$, defined by
\begin{equation}
    \label{eq:piecewiseLinearActivation}
    g(x):= \left\{
    \begin{array}{ll}
    -1, &{\rm if}~ x<-1\\
    x, &{\rm if}~ -1\le x\le 1\\
    1, &{\rm if }~ x>1
    \end{array}
    \right.
\end{equation}
It is well-known that replacing $\tanh$ with function $g$ only changes the smoothness of the model, without altering its expressivity\cite{Siegelmann:1992}.  We now give some intuitive explanation as to why these theorems hold. 

\noindent {\bf Intuition of Theorem \ref{thm:RRN1d_fails}} Consider the following parameter setting of ${\cal F}_{\rm RNN}$ for example:
$
W=\left[
\begin{array}{cc}
W_1 ,& W_2
\end{array}
\right], ~
 b=0.
$
In this setting, when feeding the $F1B(n)$ process to Vanilla RNN, the state $S_t$ evolves according to
\begin{equation}
\label{eq:rnn-F1B-stateupdate}
S_t=\tanh \left(US_{t-1}+ W_1X_t^{\rm I}+W_2X_t^{\rm F}\right)
\end{equation}
%
%
To intuitively analyze what is required for the model to pass the test, we divide our discussion into the following phases.

\noindent \underline{\em Noise Suppression ($t<L$):} In this phase,  the desired flagged information bit has not arrived and all incoming information is noise. Then the network should suppress its previous state value. 
This would require a small value of $U$.

\noindent \underline{\em Feature Loading  ($t=L$):} This is the time when the flagged information bit arrives. The network then needs to pick up the information and load it into its state. At this point, the network demands $W_1$ to be reasonably large so that the flagged bit gets amplified sufficiently when it is blended in state $S_t$. Only so can the selected feature have sufficient contrast against the background noise in the state.

\noindent \underline{\em Feature Memorization ($t>L$):} In this phase, the selected feature already resides in the state. The network must then protect the selected feature against the incoming noise.
To achieve this,  it desires  $W_1$ to be small, so that the incoming noise is sufficiently attenuated. At the same time, it desires $U$ to be large, so as to maintain a strong amplitude of the selected feature.

At this end, we see a conflict in the requirements of $U$ and $W_1$.  Such a conflict is arguably fundamental since it reflects the
incompatibility between feature selection and memorization.  Not only arising in the F1B test, such a conflict is expected to exist in much broader context of sequence learning using RNNs.

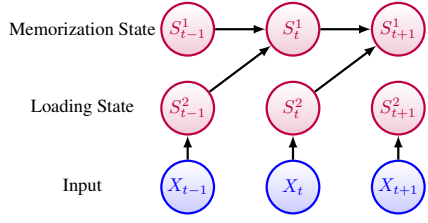
\begin{figure}
    \centering
    \scalebox{0.7}{
    \begin{tikzpicture}[-latex ,auto ,node distance =2 cm and 2 cm ,on grid,
very thick,
state/.style ={ circle ,top color =white , bottom color = processblue!20 ,
draw, processblue , text=blue , minimum width =0.05cm},
box/.style ={rectangle ,top color =white , bottom color = processblue!20 ,
draw, processblue , text=blue , minimum width =0.3cm , minimum height = 0.3cm},
neuron/.style ={rectangle ,top color =white , bottom color = red!20 ,
draw, red , text=red , minimum width =9.5cm , minimum height = 6cm, rounded corners=.5cm, opacity=0.5},
triangle/.style = {top color =white , bottom color = processblue!20 ,
draw, processblue , text=blue, regular polygon, regular polygon sides=3, minimum size=0.5cm, draw },
node rotated/.style = {rotate=270},
    border rotated/.style = {shape border rotate=270},
biasedActivationNeuron/.style ={regular polygon, regular polygon sides=5, top color =white , 
bottom color = red!35 , draw, red , text=red , 
minimum width =1.2cm},
]

\tikzset{every node/.style={inner sep=-2pt}}

\tikzstyle{add}=[state];
\tikzstyle{copy}=[state];
\tikzstyle{product}=[box];
\tikzstyle{stack}=[state, top color=red, bottom color=red];
\tikzstyle{activate}=[triangle];
\tikzstyle{complement}=[circle split, rotate=45, draw, orange!50!black, top color=orange, bottom color=white];
\tikzstyle{stochasticMap}=[state, top color =white , bottom color = purple!20, draw, purple , text=purple , minimum width =1cm];
\tikzstyle{function}=[neuron, rounded corners=.1cm, opacity=1, top color =white , bottom color = red!35 , draw, red , text=red , 
minimum width =0.8cm, minimum height=0.8cm];


\tikzstyle{dead add}=[state, top color =white , bottom color = white, draw, processblue!40!white , text=processblue!50!white, opacity=0.5];
\tikzstyle{dead copy}=[state, top color =white , bottom color = white, draw, processblue!40!white , text=processblue!50!white, opacity=0.5];
\tikzstyle{dead product}=[box, top color =white , bottom color = white, draw, processblue!40!white , text=processblue!50!white, opacity=0.5];
\tikzstyle{dead stack}=[state, top color=red!20!white, bottom color=red!20!white, draw, processblue!40!white , text=processblue!50!white, opacity=0.5];
\tikzstyle{dead activate}=[triangle, top color =white , bottom color = white, draw, processblue!40!white , text=processblue!50!white, opacity=0.5];
\tikzstyle{dead biasedActivationNeuron}=[biasedActivationNeuron, top color =white , bottom color = white, draw, red!20!white , text=red!50!white, opacity=0.5];
\tikzstyle{dead complement}=[complement, top color =white , bottom color = white, draw, orange!40!white , text=orange!50!white, opacity=0.5];

\tikzstyle{xNode}=[state, top color =white , bottom color = blue!20, draw, blue , text=blue , minimum width =1cm];


\node[](origin){Memorization State};
\node[below = 1.5cm of origin](origin1){Loading State};
\node[below = 1.5cm of origin1](origin2){Input};

\node[stochasticMap, right= 2cm of origin](s1_past){$S_{t-1}^1$};
\node[stochasticMap, below=1.5cm of s1_past](s2_past){$S_{t-1}^2$};
\node[stochasticMap, right=2cm of s1_past](s1_present){$S_{t}^1$};
\node[stochasticMap, below=1.5cm of s1_present](s2_present){$S_{t}^2$};
\node[stochasticMap, right=2cm of s1_present](s1_future){$S_{t+1}^1$};
\node[stochasticMap, below=1.5cm of s1_future](s2_future){$S_{t+1}^2$};

\node[xNode, below=1.5cm of s2_past](x_past){$X_{t-1}$};
\node[xNode, below=1.5cm of s2_present](x_present){$X_{t}$};
\node[xNode, below=1.5cm of s2_future](x_future){$X_{t+1}$};

\path (s1_past) edge (s1_present);
\path (s1_present) edge (s1_future);

\path (s2_past) edge (s1_present);
\path (s2_present) edge (s1_future);

\path (x_past) edge (s2_past);
\path (x_present) edge (s2_present);
\path (x_future) edge (s2_future);

%
%
%
%
%
%
%
%
%
%
%
%
%

\end{tikzpicture}}
    \caption{Dependency of variables in Vanilla RNN ($K=2$) with constructed parameters}
    \label{fig:rnn2d_dependency}
\end{figure}

\noindent{\bf Intuition of Theorem \ref{thm:RNN2d_pass}} 
At state dimension $K=2$, we express $S_t$ as $[S_t^1, S_t^2]^{\rm T}$. To make Vanilla RNN pass the test, the key insight is that the two states $S_t^1$ and $S_t^2$ need to assume separate functions, specifically, one responsible for memorizing the feature (``memorization state'') and the other responsible for selecting the feature (``loading state''). We now illustrate a construction of the Vanilla RNN parameters in which $S_1^t$ is chosen as the memorization state and $S_2^t$ as the loading state. 

For simplicity, we consider the cases where the flag variable $X_t^{\rm F}$ takes values in $\{0, 1\}$, instead of in $\{\pm 1\}$. We note that this modification does not change the nature of the problem (see Lemma 2 in Supplementary Materials). We also consider the case where the activation function of state update is the piece-wise linear function $g$. 

Now consider the following construction. 
\[
U:= 
\left[
\begin{array}{cc}
1     & b_{1}  \\
0 & 0 
\end{array}
\right], 
W:= 
\left[
\begin{array}{cc}
0     & 0  \\
W_{21} & -b_{2} 
\end{array}
\right] 
\]
The state-update equation then becomes 
\begin{eqnarray*}
S_{t}^{1}& = & g\left(S_{t-1}^{1}+b_{1}S_{t-1}^{2}+b_{1}\right)\\
S_{t}^{2}& = &g\left(W_{21}X_{t}^{I}-b_{2}X_{t}^{F}+b_{2}\right)
\end{eqnarray*}
This induces a dependency structure of the state and input variables as shown in Figure \ref{fig:rnn2d_dependency}. For convenience, the pre-activation term in the 
$S_t^1$ equation will be denoted by $h(S_{t-1}^1, S_{t-1}^2)$. That is,
\[
h(S_{t-1}^1, S_{t-1}^2):= S_{t-1}^1+b_1S_{t-1}^2 + b_1.
\]
We also impose the following conditions on the parameters: $W_{21} \in (0,1)$,  $b_2 <0$, $W_{21}X_{t}^{I}+b_{2}<-1$, $b_1\neq 0$, and $b_1\pm b_1 W_{21} \in (-1, 1)$.  

It is possible to verify that under such a setting, four distinct ``working mechanisms" exist. These mechanisms, their respective conditions and effects are shown in Table \ref{tab:rnn2d_mechanisms}. 


\begin{table*}[ht!]
    \caption{Working mechanisms of Vanilla RNN ($K=2$) with constructed parameters}
    \centering
    \begin{tabular}{|c|p{4cm}|p{2.9cm}|p{6.3cm}|}
    \hline
    Mechanism  & Condition  &  Effect & Comments \\
    \hline
    Load Emptying & $X^{\rm F}_t=0$ & $S_t^2=-1$ & Loading state is set to clean background (-1)\\
    \hline
    Feature Loading & $X^{\rm F}_t=1$ &$S_{t}^{2}=W_{21}X_{t}^{I}$ & Feature $X_t^{\rm I}$ is loaded into loading state \\
    \hline
    Memorization & $S_{t-1}^1\in (-1, 1)$, $S_{t-1}^2=-1$ & $S_t^1=S_{t-1}^1$ & Memorization state keeps its previous value\\
    \hline
    State Mixing & 
    $h(S_{t-1}^1, S_{t-1}^2)  \in  (-1, 1)$,
    $S_{t-1}^2  >  -1$
    &$S_t^1=h(S_{t-1}^1, S_{t-1}^2)$ & Linear mixing of loading and memorization states and storing it in memorization state \\
    \hline
    \end{tabular}
    \label{tab:rnn2d_mechanisms}
\end{table*}

Let the state be initialized as $S_{0}^{1}=0$ and $S_{0}^{2}=-1$. The dynamics of Vanilla RNN may then be divided into four phases, each controlled by exactly two mechanisms. 

















\noindent \underline{\em Noise Suppression} ($t < L$):  In this phase, the ``load emptying'' and ``memorization" mechanisms are invoked, where the former keeps the loading state $S_t^2$ clean, with value $-1$,  and the latter assures that the memorization state $S_t^1$ keeps its initial value $0$.


\noindent \underline{\em Feature Loading} ($t = L$): In this phase, ``feature loading'' is invoked, the feature $X_L^{\rm I}$ is then loaded into the loading state $S_L^2$. The ``memorization'' mechanism remains in effect, making the memorization state $S_L^{1}$ stay at $0$.


\noindent \underline{\em Feature Transfer} ($t = L+1$): In this phase, ``state mixing'' is invoked. This makes the memorization state $S_{L+1}^1$ switch to a linear mixing of the previous loading and memorization states, causing the feature contained in the previous loading state $S_L^2$ transferred to the memorization state $S_{L+1}^1$. At this time, ``load emptying" is again activated, resetting the loading state $S_{L+1}^2$ to  $-1$.


\noindent \underline{\em Feature Memorization} ($t > L+1$): In this phase, ``load emptying'' remains active, making the loading state $S_t^{2}$ stay at $-1$. At the same time, ``memorization'' is again activated, allowing the memorization state $S_t^1$ to maintain its previous value, thereby keeping the feature until the end. 

For $\tanh$-activated Vanilla RNN with $K=2$, feature selection and memorization appear to work in a similar manner, in the sense that the two states need to execute different tasks and that similar working mechanisms are involved. However, in that case, the four mechanisms are less sharply defined, and in each phase the state update is {\em dominated}, rather than {\em completely controlled}, by its two mechanisms.

\subsection{Gated RNN Models}

 \begin{thm}
 \label{thm:GRU_d1}
At state dimension $K=1$, ${\cal F}_{\rm LSTM}$, ${\cal F}_{\rm GRU}$ and ${\cal F}_{\rm PRU}$ all pass the F1B test.
 \end{thm}
 
The theorem is proved by constructing specific parameters for each of the models and for any given $n$. Under those parameters, it is possible to track the mean $\mu^+_t$ and variance ${\sigma^2_t}^+$ of state $s_t$ for the positive sample paths, and likewise
$\mu^-_t$ and ${\sigma^2_t}^-$
for the negative sample paths.
It can then be shown that the chosen parameters will make $\mu^+=-\mu^-$ and ${\sigma^2_t}^+={\sigma^2_t}^-$,  and the ratio $\sqrt{{\sigma^2_n}^+}/\mu^+_n$ can be made arbitrarily small. Then if one uses the threshold $0$ on $S_n$ to classify the positive and the negative sample paths, the error probability can be made arbitrarily close to $0$ using the Chebyshev Inequality\cite{feller2008introduction}. 
Then via a simple lemma (Lemma 3 in Supplementary Materials), arbitrary small error probability in fact implies zero error probability. 

Using GRU as an example, we now explain the construction of the model parameters and give some insights into how such a construction allows both feature selection and feature memorization.

In GRU, consider its parameter in the following form:
{\small
\begin{eqnarray*}
& & W^{\rm r} := [ 0 , 0 ]; W^{\rm z} := [ 0 , a ]; W^{\widetilde{\rm s}}: = [ b , 0 ] \\
& & U^{\rm r},U^{\rm z},U^{\widetilde{\rm s}}:= 0; b^{\rm r}, b^{\rm z}, b^{\widetilde{\rm s}}:=0
\end{eqnarray*}
}
Let $A := \sigma(a), B := \tanh(b)$.
Then the gated combination equation (\ref{eq:GRU_gatedCombine}) becomes
{
\begin{eqnarray}
S_t & =AS_{t-1}+ (1-A)B X_t^{\rm I}, & {\rm if}~ t\neq L \label{eq:GRU_analysisNotL}\\
S_t &=(1-A)S_{t-1}+ AB X_t^{\rm I}, & {\rm if}~ t = L \label{eq:GRU_analysisL}
\end{eqnarray}
}
In above, $B$ is always bundled with  $X_t^{\rm I}$, serving merely as a global constant that scales the inputs and having no impact on the problem nature. We then take $B=1$.

Observe that for both $t=L$ and $t\neq L$, the new state is a convex combination of the old state and the current input, however the weights are swapped between the two cases.  This contrasts the case of Vanilla RNN (with $K=1$), in which the same linear combination formula applies to all $t$.

This phenomenon of swapped weights can be precisely attributed to the gating mechanism in GRU, namely, that the gate variable $z_t$ in (\ref{eq:GRU_gatedCombine}) is computed dynamically from the previous state and the current input.  To analyze the effect of gating,  we discuss the consequence of (\ref{eq:GRU_analysisNotL}) and (\ref{eq:GRU_analysisL}) next.

\noindent \underline{\em Noise Suppression ($t<L$):} In this phase, (\ref{eq:GRU_analysisNotL}) applies and the network must suppress the content of the state, since it contains completely noise. This can be achieved by making $A$ either close to $0$ or  close to $1$. When $A$ is near $0$, it suppresses the noise by directly attenuating the state. When $A$ is near $1$, it suppresses the noise by attenuating the input.

\noindent \underline{\em Feature Loading ($t=L$):} At this time, (\ref{eq:GRU_analysisL}) applies. In this scenario, the network must amplify the incoming $X_L^{\rm I}$ and at the same time attenuate the old state. Only so will the desired information bit be superimposed on the noisy state with sufficient contrast. It is easy to see that both objectives can be achieved by choosing $A$ close to $1$.

\noindent \underline{\em Feature Memorization ($t>L$):} In this phase, (\ref{eq:GRU_analysisNotL}) applies again. At this time, the state contains the selected feature, and the network must sufficiently amplify the state and suppress the incoming noise.  These two objectives can be achieved simultaneously by choosing $A$ close to $1$.

Summarizing the requirements of $A$ in the three phases, we see that when choosing $A$ close to $1$, both feature selection and feature memorization can be achieved. 
It is in fact precisely due to gating that GRU is able to dynamically control what input to select and how to scale the state.

\section{Experiments}
\label{sec:exp}

\subsection{Vanilla RNN: $K=1$}

\begin{figure*}[t]\centering
\vspace{-.1cm}
\begin{tabular}{@{\hspace{0.0cm}}c@{\hspace{0.0cm}}c@{\hspace{0.0cm}}c@{\hspace{0.0cm}}c@{\hspace{0.0cm}}c@{\hspace{0.0cm}}c@{\hspace{0.0cm}}c}
\includegraphics[width=.17\textwidth]{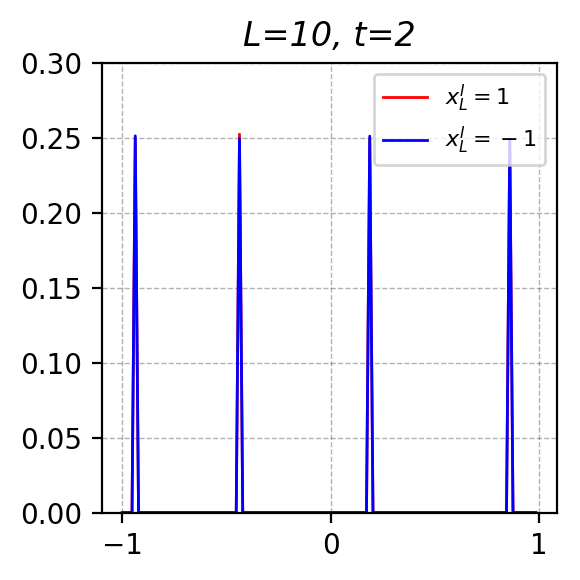}&
\includegraphics[width=.17\textwidth]{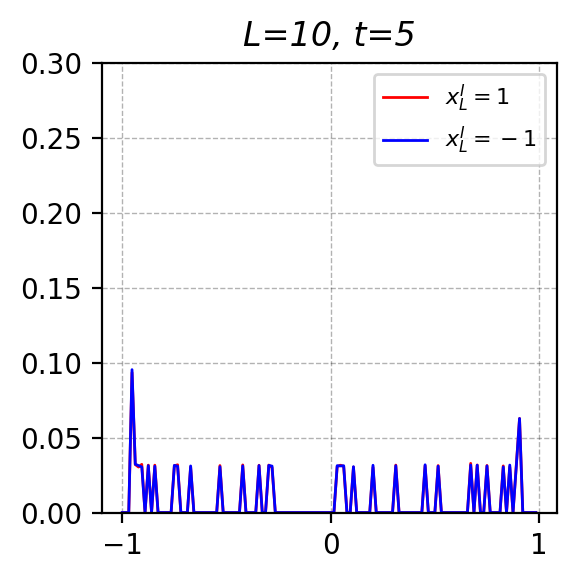}&
\includegraphics[width=.17\textwidth]{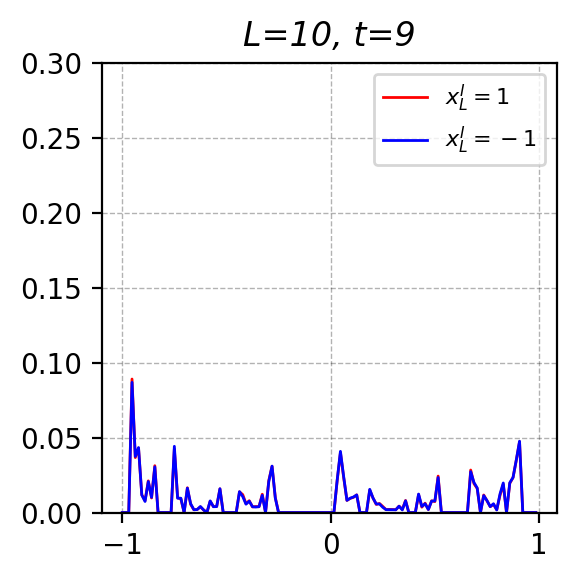}&
\includegraphics[width=.17\textwidth]{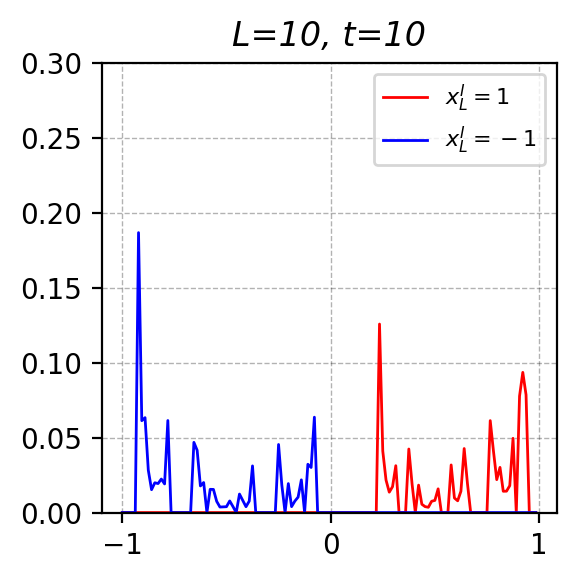}&
\includegraphics[width=.17\textwidth]{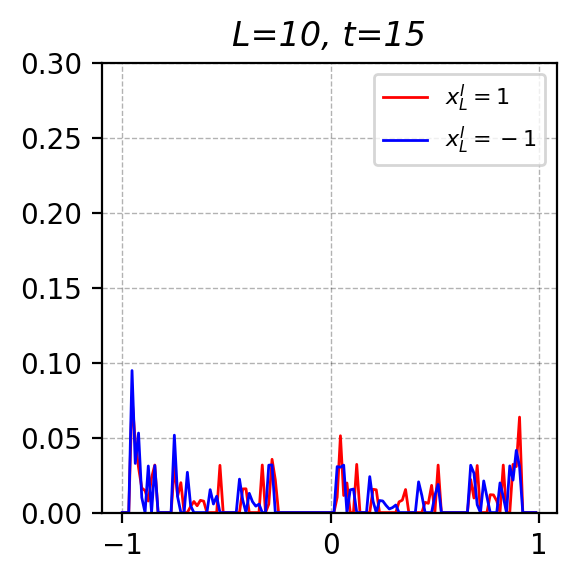}&
\includegraphics[width=.17\textwidth]{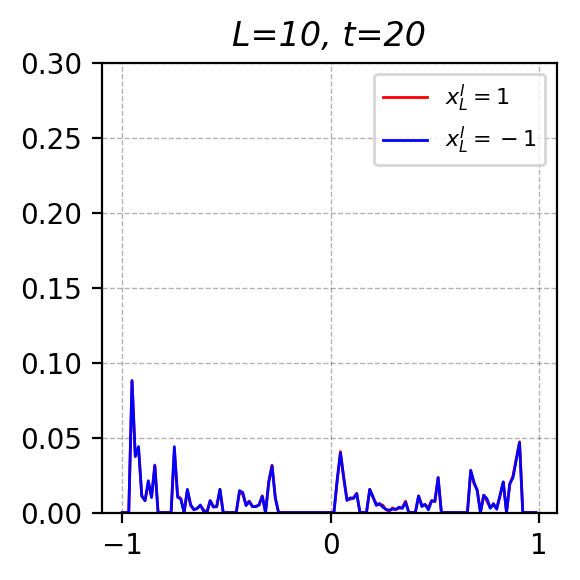}\\
\includegraphics[width=.17\textwidth]{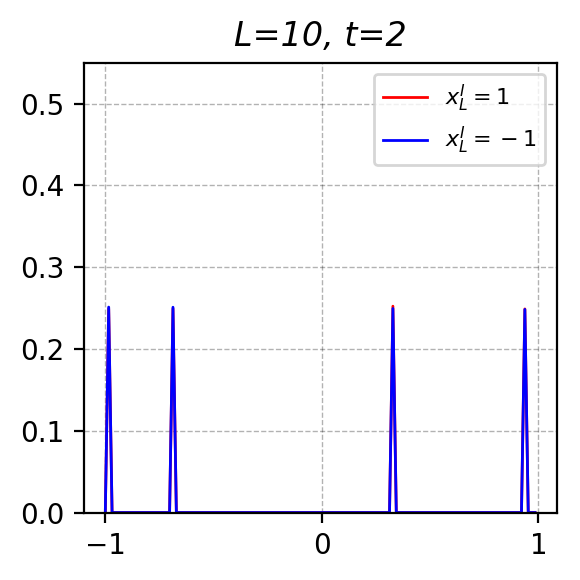}&
\includegraphics[width=.17\textwidth]{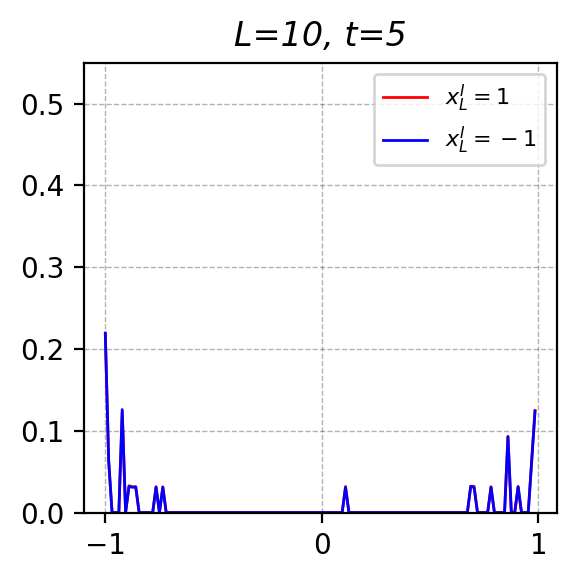}&
\includegraphics[width=.17\textwidth]{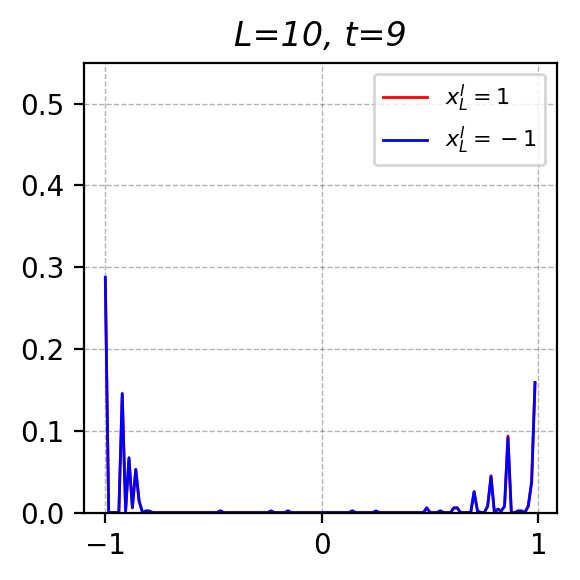}&
\includegraphics[width=.17\textwidth]{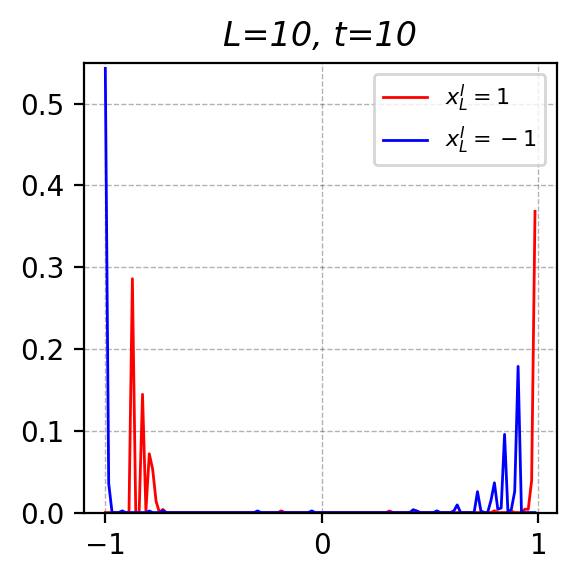}&
\includegraphics[width=.17\textwidth]{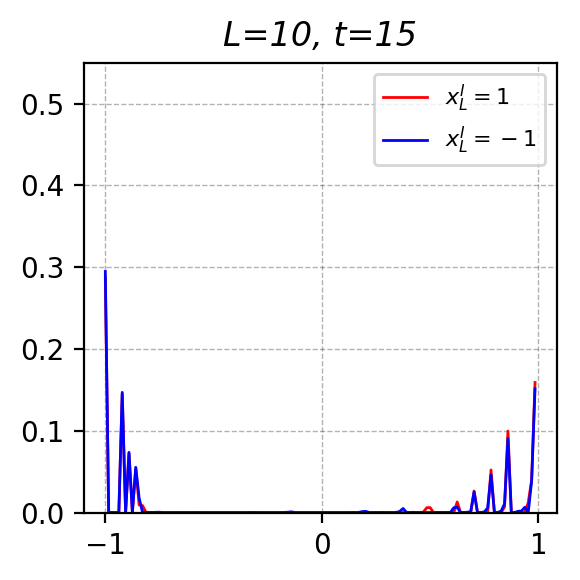}&
\includegraphics[width=.17\textwidth]{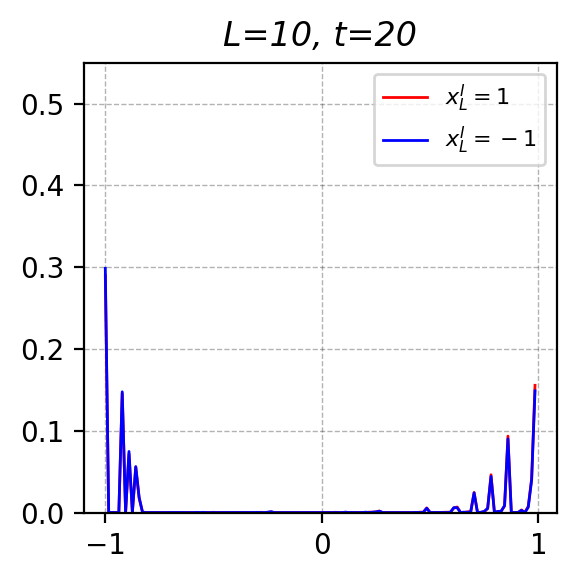}
\end{tabular}
\vspace{-.3cm}
\caption{\label{fig:RNN1d}Distribution of state $S_t$, for various $t$,  in Vanilla RNN with state dimension 1. Input sequence length $n=20$. 
Flag location $L=10$. Red curves: state distributions of positive sample paths. Blue curves: state distributions of negative sample paths.  
Top row: $U=0.8$, $W_1=0.9$, and $W_2=0.1$. Bottom row: $U=2.0$, $W_1 = 0.7$, $W_2 = 0.1$. The parameters are chosen in a way such that the evolution of the distributions is not too fast, allowing for better visualization.}
\vspace{-.3cm}
\end{figure*}

We have performed extensive computer search in the parameter space of ${\cal F}_{\rm RNN}$, and all checked parameter settings fail the F1B test. Specifically, we apply each checked parameter setting to the Vanilla RNN model and feed $10,000$ randomly generated positive sample paths of length $1000$  to the model. This Monte-Carlo simulation then give rise to the final state distribution for the positive sample paths. Similarly, we obtain the final state distribution for the negative sample paths. For each examined parameter, we observe that the two distributions significantly overlap, suggesting a failure of the F1B test.

The results in Figure \ref{fig:RNN1d} are obtained by similar Monte-Carlo simulations as above. The figure shows some typical behaviours of Vanilla RNN with state dimension $K=1$. The first row of the figure corresponds to a case where the value of $U$ is relatively small. In this case, in the noise-suppression phase, the model is able to suppress the noise in the state, and we see the state values move towards the centre. This then allows the success in feature loading, where the positive distribution is separated from the negative distribution. However, such a parameter setting makes the network lack memorization capability. In the feature-memorization phase, we see the two distributions quickly overlap, and can not be separated in the end. That is, the memory of the feature is lost. The second row of the figure corresponds to a case where $U$ is relatively large. In this case, the model is unable to compress the state values in the noise-suppression phase. As a consequence, when the feature is loaded into the state, the positive and the negative distributions can not be separated by a linear classifier. That is, feature selection fails. Then in the feature-memorization phase, the large value of $U$ keeps magnifying the state values in order to keep its memory. As noise is continuously blended in, this effort of ``memorization''  makes the two distributions even more noisy and in the end completely overlap.

\begin{figure*}[t]\centering
\vspace{-.1cm}
\begin{tabular}{@{\hspace{0.0cm}}c@{\hspace{0.0cm}}c@{\hspace{0.0cm}}c@{\hspace{0.0cm}}c@{\hspace{0.0cm}}c@{\hspace{0.0cm}}c}
\includegraphics[width=.17\textwidth]{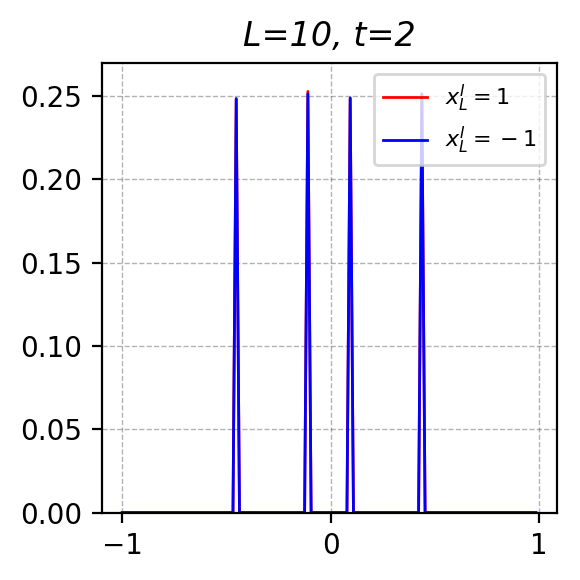}&
\includegraphics[width=.17\textwidth]{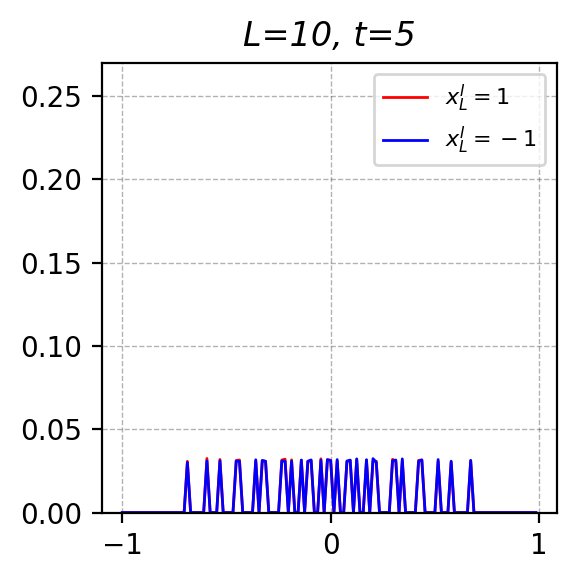}&
\includegraphics[width=.17\textwidth]{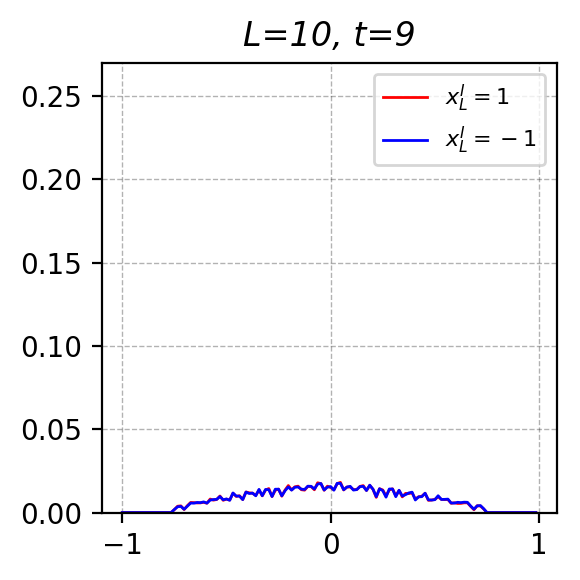}&
\includegraphics[width=.17\textwidth]{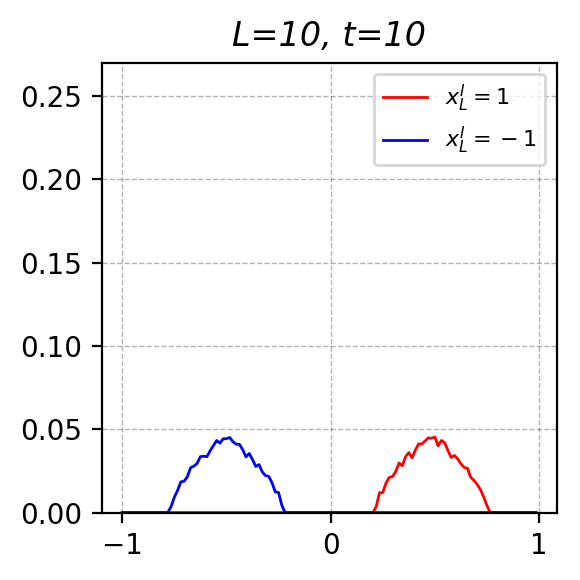}&
\includegraphics[width=.17\textwidth]{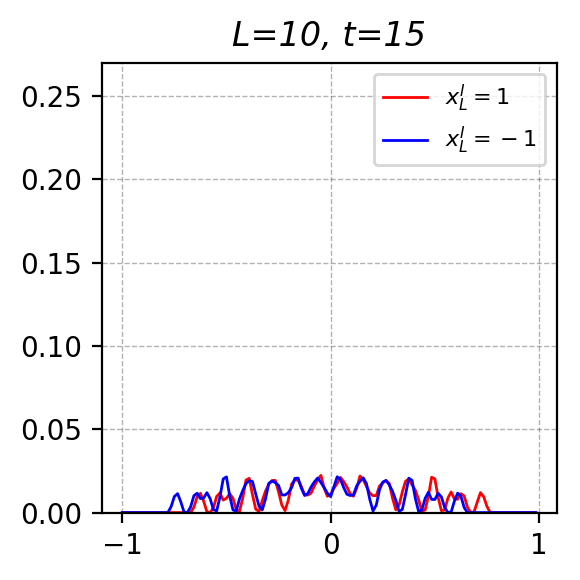}&
\includegraphics[width=.17\textwidth]{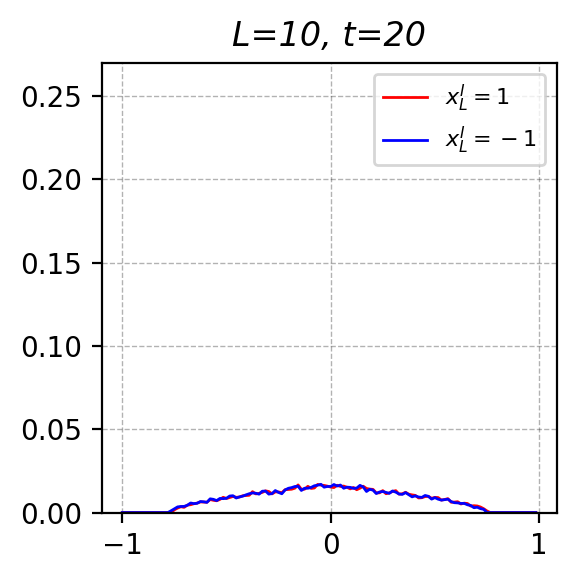}\\
\includegraphics[width=.17\textwidth]{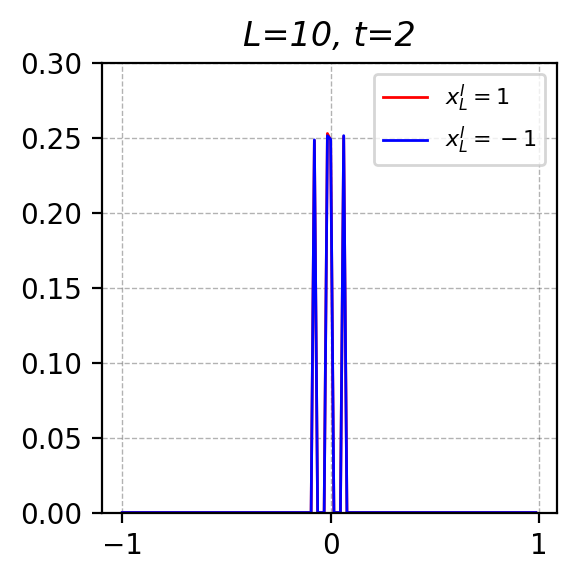}&
\includegraphics[width=.17\textwidth]{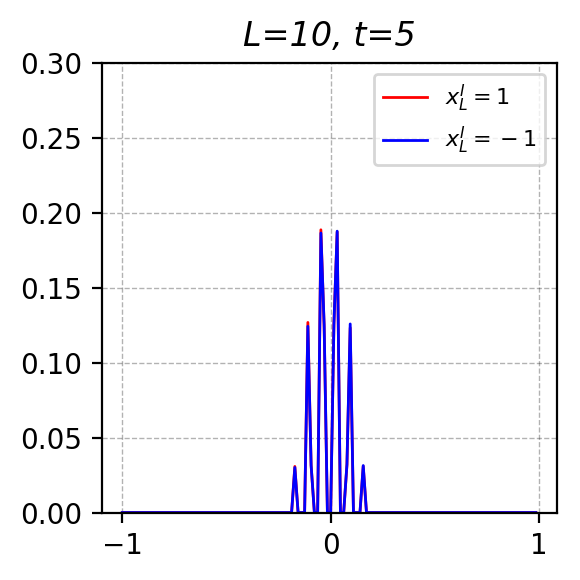}&
\includegraphics[width=.17\textwidth]{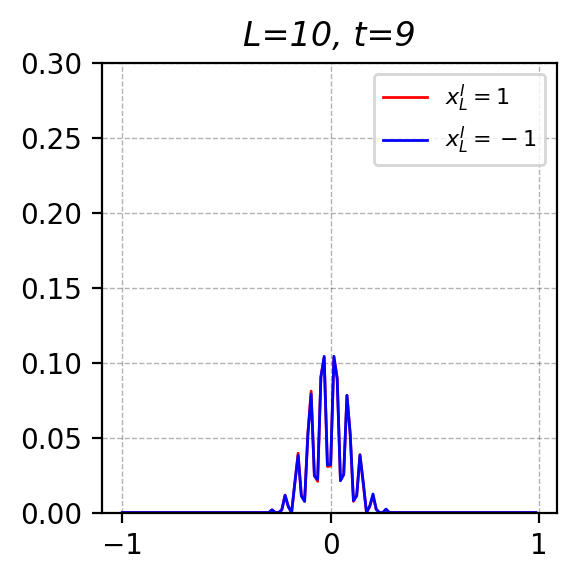}&
\includegraphics[width=.17\textwidth]{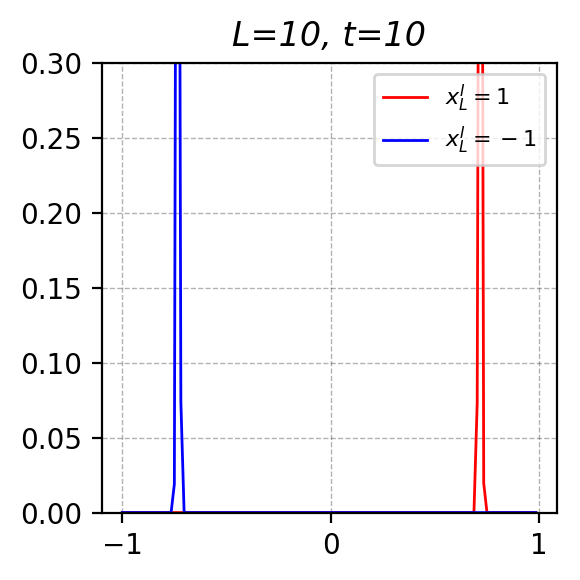}&
\includegraphics[width=.17\textwidth]{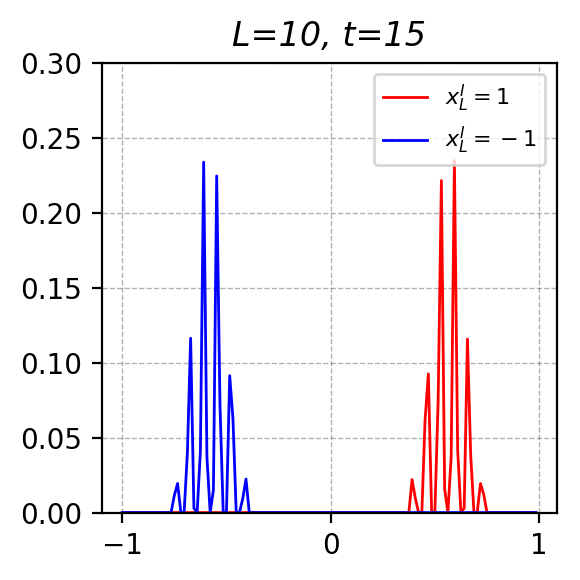}&
\includegraphics[width=.17\textwidth]{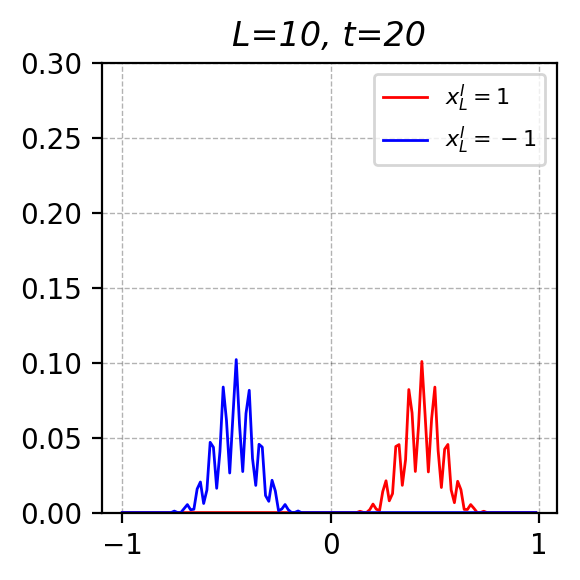}\\
\vspace{-.3cm}
\end{tabular}
\caption{\label{fig:GRU1d}Positive (red) and negative (blue) distribution of state $S_t$, at various $t$,  for GRU with state dimension 1 (the same set up as that in Figure \ref{fig:RNN1d}).
Top row: $a=0.5$, $b=1.0$ ($A$ = 0.62, $B = 0.76$). Bottom row: $a =4.0$, $b$ = 1.0 ($A$ = 0.98, $B=0,76$).}
\vspace{-.3cm}
\end{figure*}

\subsection{GRU: $K=1$}
Figure \ref{fig:GRU1d} shows some Monte-Carlo simulation results,
obtained similarly as those in Figure \ref{fig:RNN1d}, that validate Theorem \ref{thm:GRU_d1}.  The top row of the figure corresponds to a case where $A$ is chosen not close to 1.  In this case, we see that in the noise-suppression phase, the state values move away from the centre, suggesting an increased noise power in the state. This noise is however not too strong, which allows the feature to be loaded in the state reasonably well. This is indicated by the well separated positive and negative distributions in the feature-loading phase. But this value of $A$ does not support feature memorization. After the feature is loaded, the two distributions quickly smear into each other and become completely overlapping in the end. The bottom row of the figure corresponds to a case where $A$ is chosen close to 1 (as suggested by the construction). In this case, the state is well suppressed in the noise-suppression phase, allowing for perfect feature loading. The positive and negative distributions are kept well separated in the feature memorization phrase, allowing for perfect linear classification. Hence GRU passes the test.

Previous explanations on the effectiveness of gating are mostly based on gradient arguments. For example, in \cite{hochreiter1997longLSTM}, introducing gates is for ``trapping'' the gradient signal in back propagation. Recently the effectiveness of gating is also justified using a signal-propagation model based on dynamical isometry and mean field theory ~\cite{ChenPS18}. This paper demonstrates the power of gating from a different perspective, namely, that regardless of the training algorithm, gating intrinsically allows the network to dynamically adapt to the input and state configuration. This perspective, to the best of our knowledge, is made precise for the first time.

\subsection{Vanilla RNN: $K=2$}

\begin{figure*}[ht!]\centering
\vspace{-.1cm}
\begin{tabular}{@{\hspace{0.0cm}}c@{\hspace{0.0cm}}c@{\hspace{0.0cm}}c@{\hspace{0.0cm}}c@{\hspace{0.0cm}}c@{\hspace{0.0cm}}c@{\hspace{0.0cm}}c@{\hspace{0.0cm}}c}
\includegraphics[width=.143\textwidth]{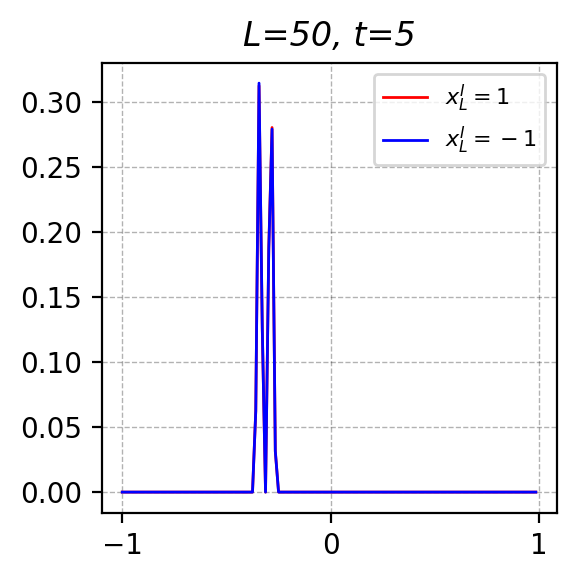}&
\includegraphics[width=.143\textwidth]{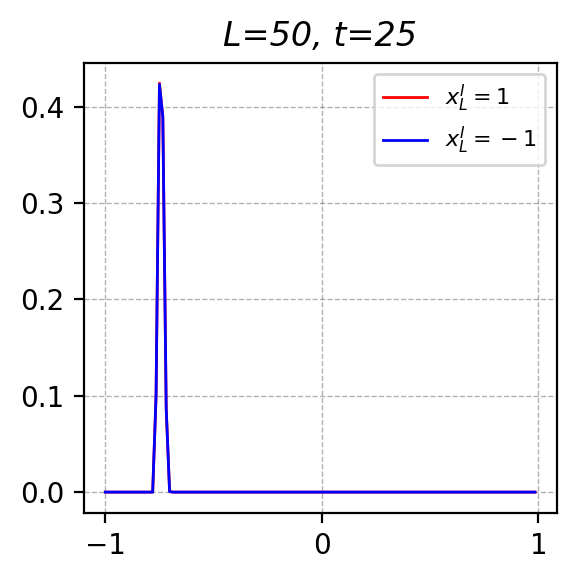}&
\includegraphics[width=.143\textwidth]{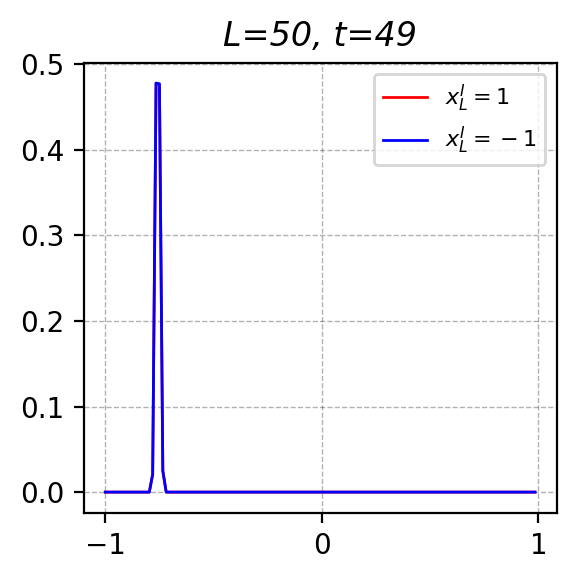}&
\includegraphics[width=.143\textwidth]{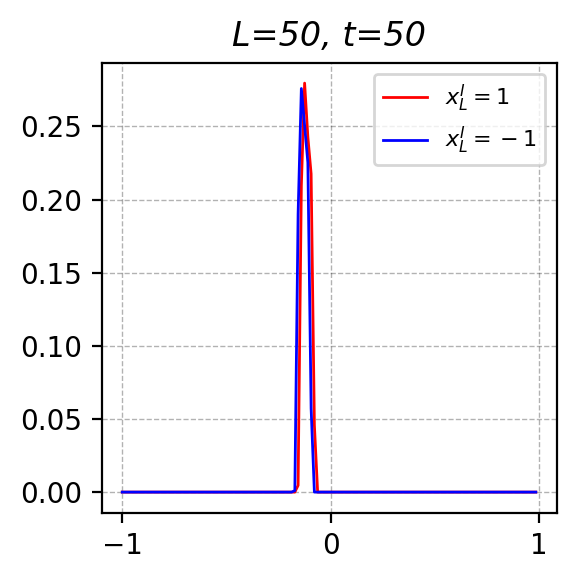}&
\includegraphics[width=.143\textwidth]{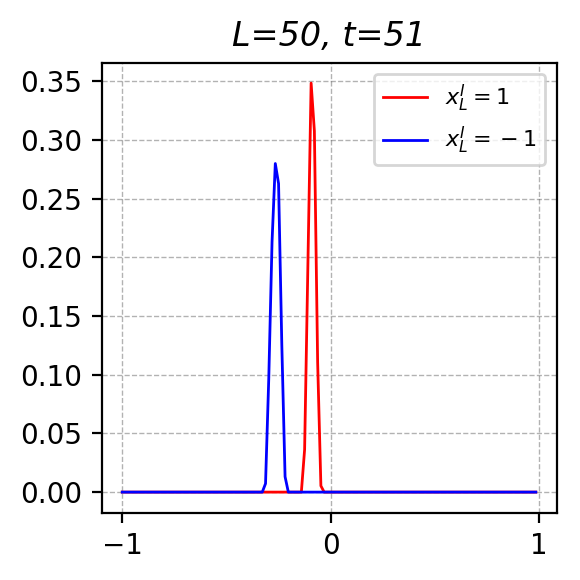}&
\includegraphics[width=.143\textwidth]{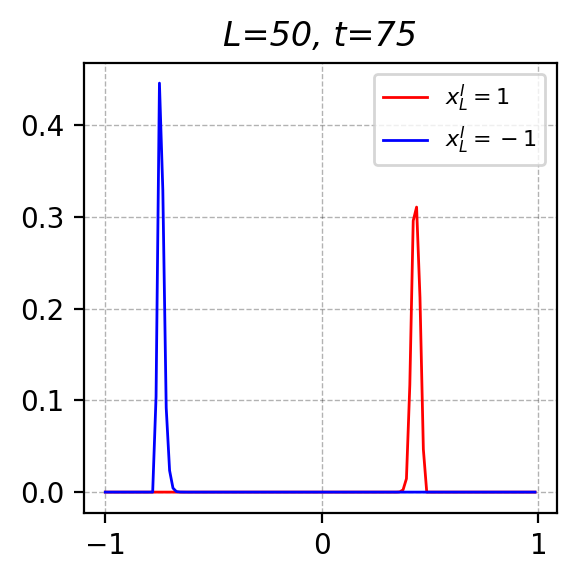}&
\includegraphics[width=.143\textwidth]{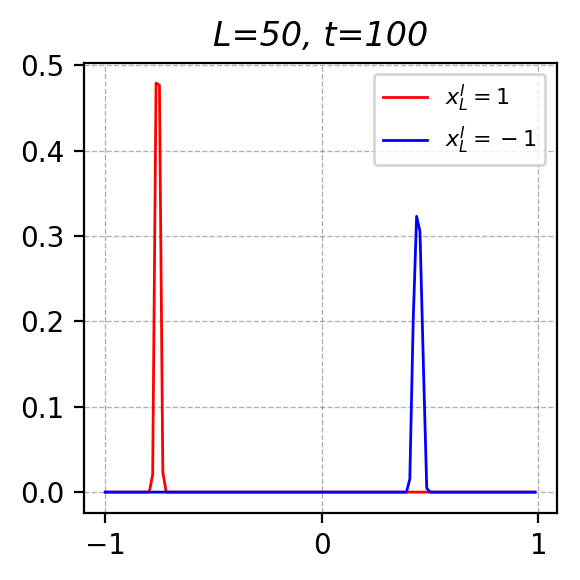}\\
\includegraphics[width=.143\textwidth]{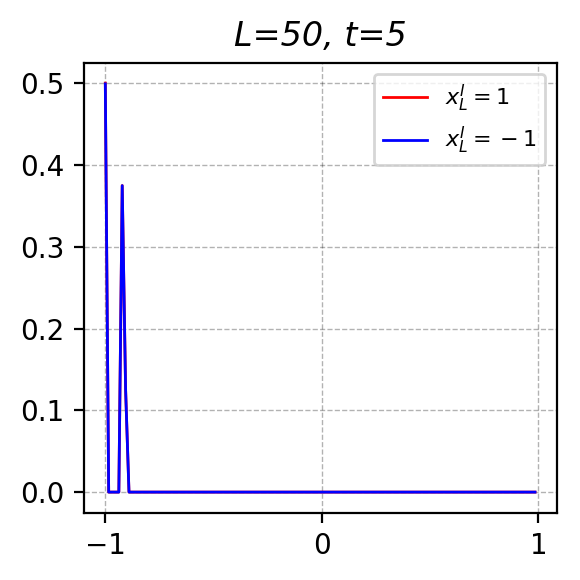}&
\includegraphics[width=.143\textwidth]{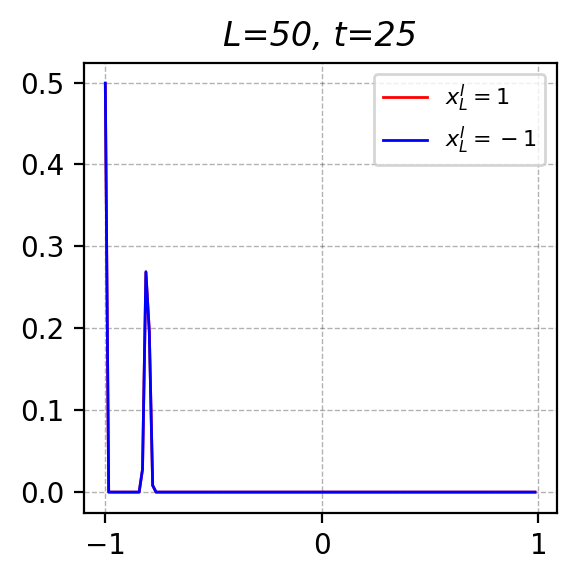}&
\includegraphics[width=.143\textwidth]{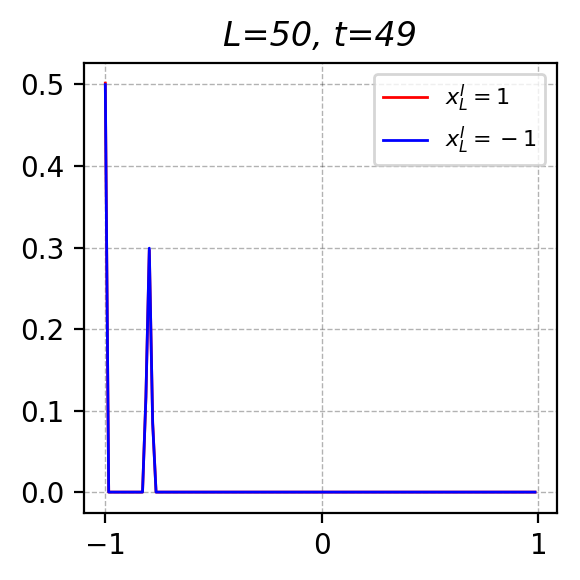}&
\includegraphics[width=.143\textwidth]{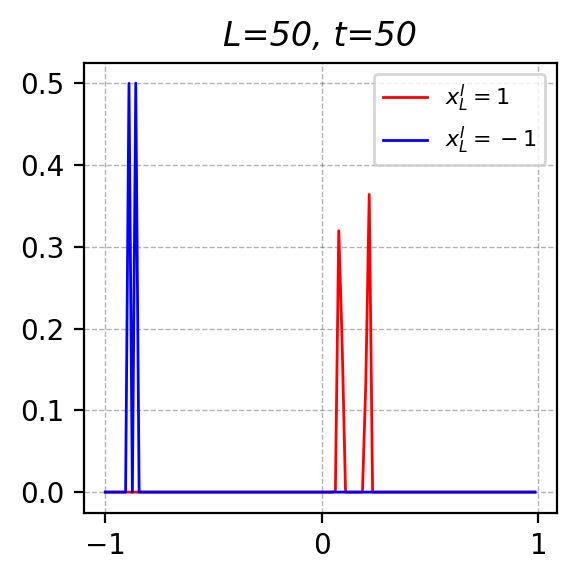}&
\includegraphics[width=.143\textwidth]{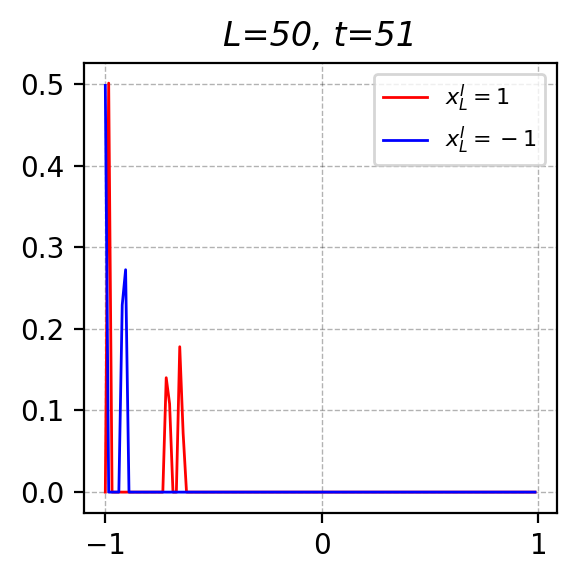}&
\includegraphics[width=.143\textwidth]{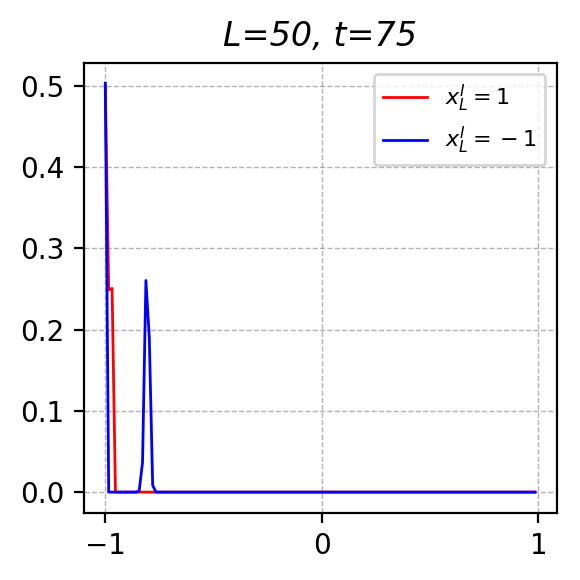}&
\includegraphics[width=.143\textwidth]{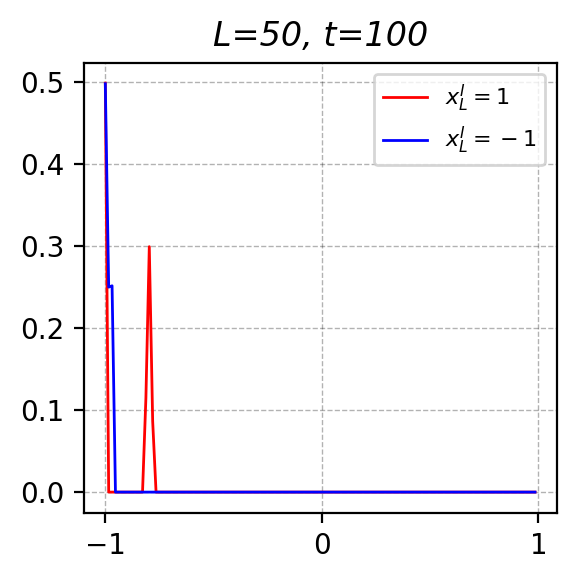}\\
\includegraphics[width=.143\textwidth]{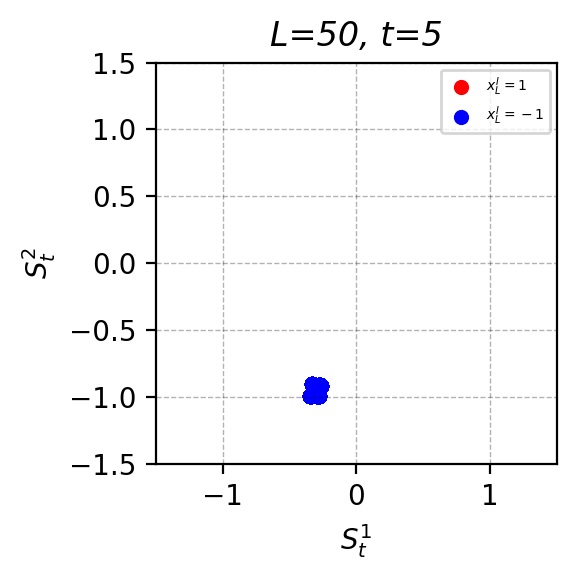}&
\includegraphics[width=.143\textwidth]{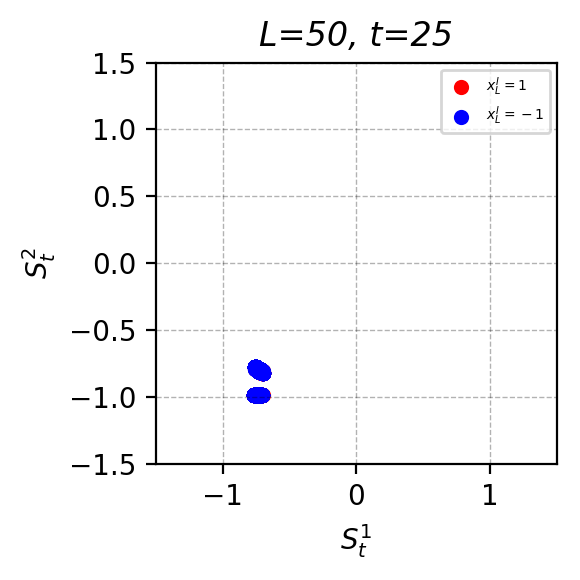}&
\includegraphics[width=.143\textwidth]{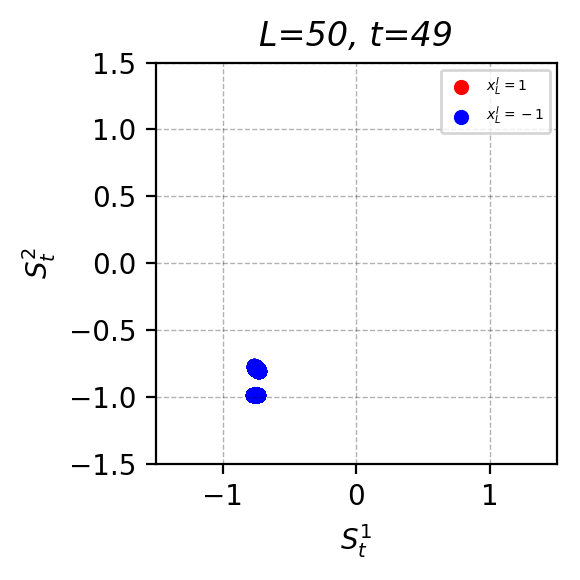}&
\includegraphics[width=.143\textwidth]{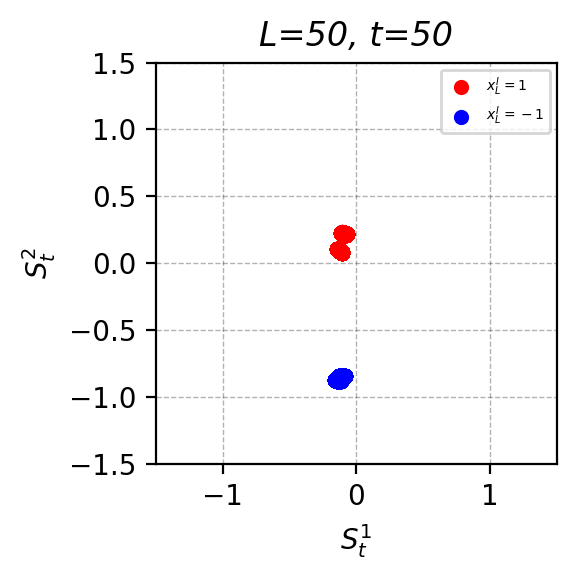}&
\includegraphics[width=.143\textwidth]{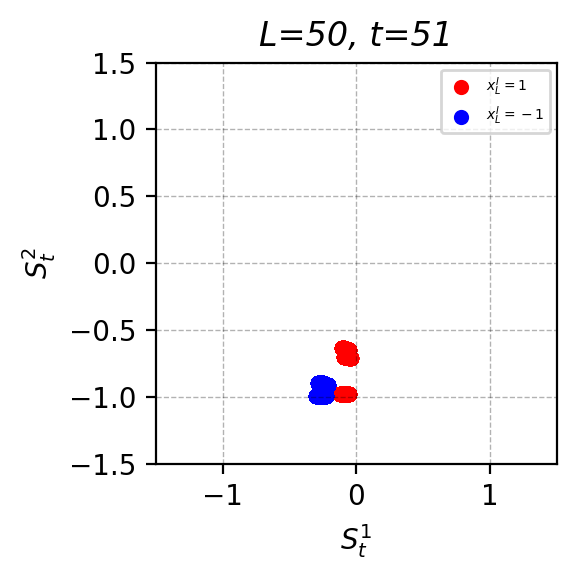}&
\includegraphics[width=.143\textwidth]{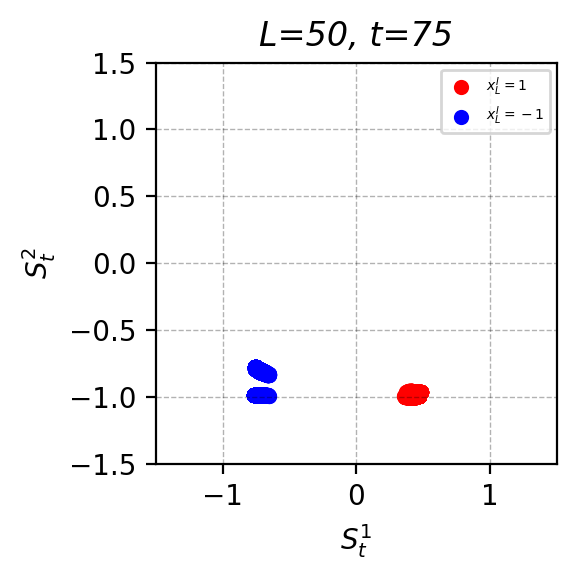}&
\includegraphics[width=.143\textwidth]{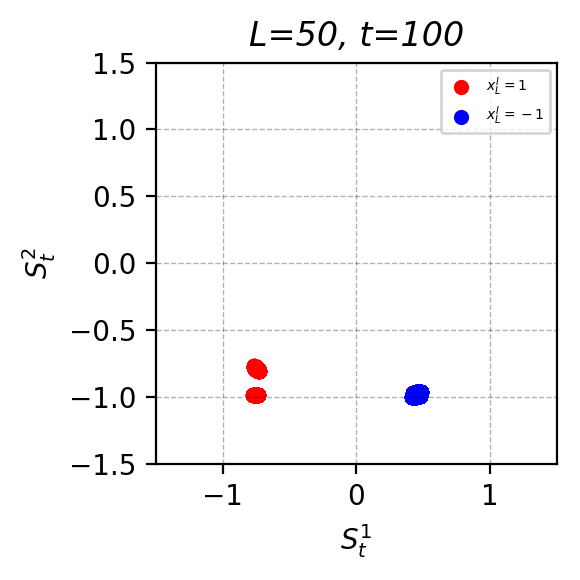}
\end{tabular}
\vspace{-.3cm}
\caption{\label{fig:RNN2d_tanh} Distributions of state $S_t^{1}$ (top row), state $S_t^{2}$ (middle row) and the scatter plots of $(S_t^{1}, S_t^{2})$ (bottom row), at various $t$, in $\tanh$-activated Vanilla RNN with state dimension $2$. Sequence length $n=100$. Flag location $L=50$. Distributions/scattering patterns for positive (resp., negative) sample paths are shown in red (resp., blue). Results are obtained using 100000 positive sample paths and 100000 negative sample paths.
Parameters: $U_{11}=-1.2$, $U_{12}=0.1938$, $U_{21}=0.8660$, $U_{22}=0.6481$,$W_{11}=0.0073$, $W_{12}=-0.3010$, $W_{21}=0.7336$, $W_{22}=1.1052$.
}
\vspace{-.3cm}
\end{figure*}

For various sequence lengths $n$, we train Vanilla RNN with state dimension $K=2$ where the activation function in the state-update equation is $\tanh$ (some results of Vanilla RNN with $K=2$ and the piece-wise linear activation function $g$ are given in Supplementary Materials). During training,  the final state $S_n$ is sent to a learnable logistic regression classifier. Mini-batched back-propagation is used to train the model. With increasing $n$, the chance of successful training (namely, training loss converging to near zero) decreases significantly, due to the well-known gradient vanishing problem of Vanilla RNN.  Figure \ref{fig:RNN2d_tanh} shows some Monte-Carlo simulation results obtained for a Vanilla RNN parameter setting learned successfully,  the shown behaviour is typical in all successfully learned models. 

The learned model in Figure \ref{fig:RNN2d_tanh} is a case where the $S_t^1$ serves as the memorization state and $S_t^2$ as the loading state. When $t<L=50$, $S_t^2$ remains at a low value, providing a clean background for the loading of the upcoming feature. When $t=L$, the feature $X_L^{\rm I}$ is loaded into the loading state $S_L^2$, where the positive and negative sample paths exhibit themselves as two separable distributions. When $t=L+1$, the feature is transferred to the  memorization state $S_{L+1}^1$, where positive sample paths are separable from the negative ones. Note that at this point, information about the feature is in fact lost in the loading state $S^2_{L+1}$ and the two distributions on the loading state are no longer separable. But the feature in the memorization state remains memorized for $t>L$, which allows the eventual classification of the positive and negative paths.


\vspace{-.3cm}
\section{Concluding Remarks}
\vspace{-.2cm}
We propose to use the ``Flagged-1-Bit'' (F1B) test to RNN's behaviour in selecting and memorizing temporal features.  This test strips off the influence of any training algorithm and reveals the models' intrinsic capability. 
Using this test, we articulate that there is in general a conflict between feature selection and feature memorization. Although the F1B test is a simplified special case, it reveals some insights concerning how RNN deals with this conflict, which might be extrapolated to a broader context of sequence learning.

Specifically with limited state dimension, with or without gate appears to be ``the great divide'' in the models' ability and behaviour to resolve such a conflict. Gating allows the network to dynamically adjust its state-update equation, thereby resolving this conflict. Without gates,  Vanilla RNN can only exploit the freedom in its state space to accommodate these conflicting objectives. In particular, we point out that  Vanilla RNN may resolve this conflict by assigning each state dimension a different function; in the special case of the F1B test and with 2-dimensional state space, one dimension serves to select feature, and the other serves to memorize it. Such an insight is expected to extend to more complex sequence learning problems, although much study is required in those settings. One potential direction is to consider a more general ``Flagged-$k$-Bit'' test, in which $k$ randomly spread flags specify the locations of $k$ desired information bits. 

This research also implies that if not subject to gradient-based training methods, Vanilla RNN, with adequate state dimensions, is not necessarily inferior to gated models. Thus fixing the gradient problem and developing new optimization techniques for Vanilla RNN remain important.




Finally, consider the spectrum of models from Vanilla RNN, to PRU, to GRU, and to LSTM. At one end of the spectrum, Vanilla RNN uses no gate and relies on increasing state dimension to resolve the feature selection-memorization conflict. At the other end,  LSTM uses a complicated gate structure to resolve the same conflict. This observation makes one wonder whether a better compromise may exploit both worlds, i.e., using a simpler gate structure and slightly increased state dimensions. Indeed, experimental evidences have shown that with simpler gate structures, GRU and PRU may outperform LSTM\cite{chung2014empirical,long2016Prototypicalru}.


\clearpage

\clearpage

\bibliography{DeepLearning}

\bibliographystyle{icml2019}

\end{document}